\documentclass[letterpaper, 10 pt, conference]{ieeeconf}  

\IEEEoverridecommandlockouts                              

\usepackage{graphicx} 
\usepackage{multirow}
\usepackage{array}
\usepackage{subfig}
\usepackage{hyperref}
\usepackage{textcomp}
\usepackage{amsmath, amssymb}

\title{\bf
LatentSLAM: unsupervised multi-sensor representation learning\\ for localization and mapping
}

\author{Ozan \c{C}atal$^{1}$, Wouter Jansen$^{2}$, Tim Verbelen$^{1}$, Bart Dhoedt$^{1}$ and Jan Steckel$^{2}$
\thanks{$^{1}$IDlab, Ghent University - imec,
        {\{firstname\}.\{lastname\}@ugent.be}}%
\thanks{$^{2}$CoSys-Lab, University of Antwerp - Flanders Make,
        {\{firstname\}.\{lastname\}@uantwerpen.be}}
\thanks{
        \textcopyright 2021 IEEE.  Personal use of this material is permitted.  Permission from IEEE must be obtained for all other uses, in any current or future media, including reprinting/republishing this material for advertising or promotional purposes, creating new collective works, for resale or redistribution to servers or lists, or reuse of any copyrighted component of this work in other works.}%
}

\begin{document}

\maketitle
\thispagestyle{empty}
\pagestyle{empty}

\begin{abstract}
Biologically inspired algorithms for simultaneous localization and mapping (SLAM) such as RatSLAM have been shown to yield effective and robust robot navigation in both indoor and outdoor environments. One drawback however is the sensitivity to perceptual aliasing due to the template matching of low-dimensional sensory templates. In this paper, we propose an unsupervised representation learning method that yields low-dimensional latent state descriptors that can be used for RatSLAM. Our method is sensor agnostic and can be applied to any sensor modality, as we illustrate for camera images, radar range-doppler maps and lidar scans. We also show how combining multiple sensors can increase the robustness, by reducing the number of false matches. We evaluate on a dataset captured with a mobile robot navigating in a warehouse-like environment, moving through different aisles with similar appearance, making it hard for the SLAM algorithms to disambiguate locations.
\end{abstract}

\section{Introduction}

Simultaneous localization and mapping (SLAM) has been a long standing challenge in robotics for the last 30 years~\cite{Cadena2016}. Most popular approaches aim to build a metric map of the environment, and then use some flavour of Bayesian filtering to estimate the robot's position on this map~\cite{Artal2015}. These approaches require a kind of observation model that maps raw sensory observations to a pose in metric space. An important outstanding challenge is then how to maintain and update this metric map over a long period of time for a dynamic environment~\cite{Shi2020}.

A different, bio-inspired method was proposed in RatSLAM~\cite{Milford2004}. The method is based on a hippocampal model of a rat's brain, where the map is represented as a topological graph. Each node in the graph is described by a robot pose ($x$, $y$, $\theta$) and a feature vector summarizing the sensor values at that scene.
This has been shown to work well for navigating a robot based on camera sensors~\cite{Milford2004}\cite{Ball2013}\cite{Gu2019}, sonar sensors~\cite{Steckel2013}, tactile sensors~\cite{Fox2012} or a combination thereof~\cite{Struckmeier2019}\cite{Jacobson2012}. Maybe the biggest advantage of RatSLAM-based approaches is their robustness against changes in the environment, such as in the case of day-night cycles~\cite{Glover2019} or lifelong SLAM ~\cite{Milford2010} which demonstrated an autonomous robot navigating in an office workspace for a 2 week period.

Despite these accomplishments, the efficiency of the system very much depends on the construction of these feature vectors, and how well these serve to disambiguate different locations~\cite{muller2014}. For example, camera images are typically downscaled and vectorized~\cite{Milford2004}, or pre-processed using LATCH features~\cite{Gu2019} or saliency maps~\cite{Yu2020}. However, recent advances in deep neural networks have shown that convolutional neural networks (CNN) can be trained for place recognition~\cite{Chen2017}, and that CNNs pretrained on for example ImageNet classification can serve as powerfull feature extractors~\cite{Kornblith2019}.

In this paper, we propose LatentSLAM, an extension to RatSLAM which uses a novel, unsupervised learning method for obtaining compact latent feature representations. Similar to RatSLAM, our method is bio-inspired and builds on the active inference framework~\cite{Friston2013life}, where living creatures are thought to minimize surprise, or equivalently the variational free energy, in order to form beliefs over the world. We show that our latent representations can be trained in a sensor-agnostic manner for multiple sensor modalities, and improve localization robustness by combining sensors. We benchmark in a highly ambiguous, warehouse-like environment, with multiple aisles between racks and shelves that very much look alike.

In short, the contribution of this paper is three-fold:
\begin{itemize}
\item We present an unsupervised learning approach for generating compact latent representations of sensor data, that can be used for RatSLAM.
\item We illustrate how this can be applied for different sensor modalities, such as camera images, radar range-doppler maps and lidar scans.
\item We provide a dataset of over 60GB of camera, lidar and radar recordings of a mobile robot in a challenging environment~\footnote{Our dataset is available at \url{https://thesmartrobot.github.io/datasets}}.
\end{itemize}

In the next section we first give an overview of the RatSLAM algorithm. In Section~\ref{sec:latent} we present LatentSLAM, our SLAM system with latent feature vectors obtained by unsupervised representation learning. Section~\ref{sec:dataset} describes our dataset, which we use in Section~\ref{sec:experiments} to benchmark our algorithm and discuss the results. Finally, Section~\ref{sec:conclusion} concludes this paper.

\section{RatSLAM}
\label{sec:ratslam}
RatSLAM is a SLAM system based on models of the hippocampus, more specifically the navigational processes in a rat's hippocampus~\cite{Milford2016}. The system consists of three components: pose cells, local view cells and an experience map~\cite{Milford2004}.

\subsubsection*{Pose Cells }
Pose cells form a Continuous Attractor Network (CAN)~\cite{Battaglia1998} configured in a wrapped around three-dimensional cube which is topologically equivalent to a torus in SE(3). The dimensions of this cube represent the pose (i.e. the $x$ and $y$ position and rotation around the Z-axis $\theta$) of a ground-based robot. A cell in the cube is active if the robot is thought to be positioned closely to the corresponding pose. After a given time the stable activation pattern of the cells will be a single cluster with the centroid corresponding with best estimate of the current pose~\cite{Milford2004,Milford2016}.

\subsubsection*{Local View Cells}
Local View Cells are organised as a list, to which a new cell is added when the robot encounters a novel scene. A single local view cell corresponds to the observation of that scene and a link is added between that cell and the centroid of the dominant activity packet in the pose cells at that time, a process equivalent to Hebbian learning~\cite{Choe2013}. When the robot re-encounters the scene of a given local view cell  activation energy will be injected in the corresponding pose cells. This way, subsequent observations that are in line with predicted motion will enforce the energy packets in the pose cell network, and therefore converge to a single peak through the local attractor dynamics of the CAN.
Every time the robot makes a camera observation, a template is generated from the image by cropping, subsampling and converting it to grey-scale. This template is then compared to the stored visual templates of the local view cells by calculating the sum of absolute differences (SAD). If the template matches a stored template -- i.e. the SAD is above a certain threshold $\delta_{match}$-- the corresponding local view cell is activated. Otherwise a new local view cell is added to the list~\cite{Milford2004,Milford2016,Ball2013}.

\subsubsection*{Experience Map}
The final component of RatSLAM is the experience map. As pose cells only represent a finite area, a single cell can and should represent multiple physical places due to the wrapping behaviour of the cube's edges. The Experience map offers an unique estimate of the robot's global pose by combining the information of the pose cells and the local view cells. The experience map itself is a graph where each node represents a certain state in the pose cells and local view cells. If a the state of these components does not match with any already existing node in the graph a new experience node is created and added to the graph. On a transition between experiences a link is formed between the previous and current experience~\cite{Milford2004,Milford2016}.

\section{LatentSLAM}
\label{sec:latent}
\begin{figure}
    \centering
    \includegraphics[width=0.25\textwidth]{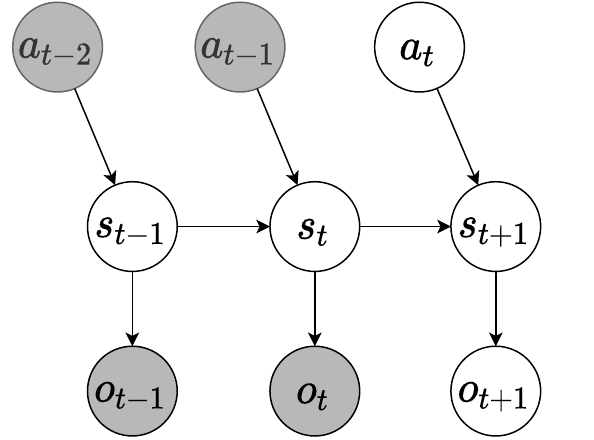}
    \caption{Schematic overview of the generative model underlying the latent state representation of our system. A transition between states $s$ is initiated by some action $a$. Observations $o$ can be fully predicted from the model states.}
    \label{fig:generative model}
\end{figure}
A common shortcoming of RatSLAM is its sensitivity to perceptual aliasing~\cite{muller2014}, in part due to the reliance on an engineered visual processing pipeline. We aim to reduce the effects of perceptual aliasing by replacing the perception module by a learned dynamics model. We create a generative model that is able to encode sensory observations into a latent code that can be used as a replacement to the visual input of the RatSLAM system. The model is able to capture environment dynamics to enable the distinction between similar observations with a different history.

\begin{figure}
	\centering
    \includegraphics[width=0.45\textwidth]{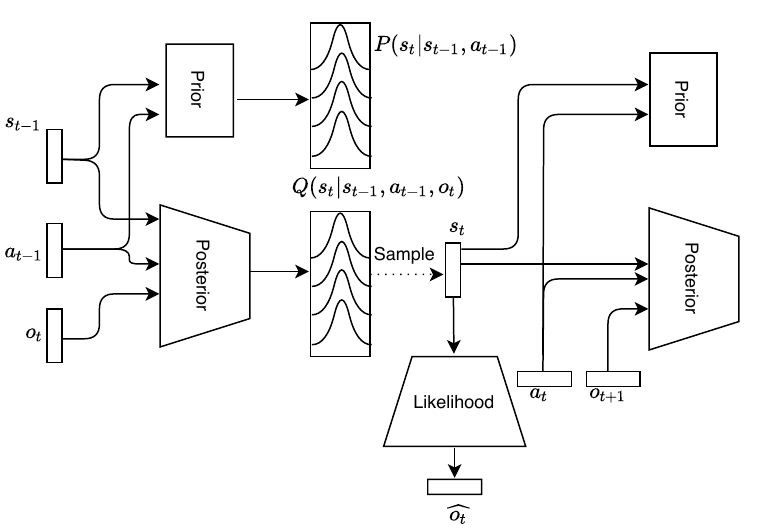}
    \caption{Schematic overview of how the different neural networks cooperate. At a given timestep the prior network generates a prior estimate on the state distribution from the previous state estimate and the previously taken action. Similarly, after observation the posterior network generates a posterior estimate on the state distribution from the previous state-action pair and the current observation. The posterior distribution is then sampled to create a new state estimate. 
    }
    \label{fig:model}
\end{figure}

Similar to RatSLAM we draw inspiration from neuroscience to create a biologically plausible approach to generative modeling based on the free energy principle~\cite{Friston2013life}, which states that organisms actively try to minimize the suprisal of incoming observations. Concretely this means we build a generative model over actions, observations and belief states as shown in Figure~\ref{fig:generative model}. Greyed out nodes represent observed variables, while white nodes need to be inferred. The model incorporates actions to generate a state estimate that will explain the corresponding observation, as described by the joint distribution over observations, states and actions in~Equation~\ref{eq:joint}.
\begin{equation}
\begin{split}
    P(\tilde{\boldsymbol{o}}, \tilde{\boldsymbol{s}}, \tilde{\boldsymbol{a}}) &=
     P(\boldsymbol{s}_0) \prod_{t=1}^{T} P(\boldsymbol{o}_t | \boldsymbol{s}_t) P(\boldsymbol{s}_t | \boldsymbol{s}_{t-1}, \boldsymbol{a}_{t-1}) P(\boldsymbol{a}_{t-1})
    \label{eq:joint}
\end{split}
\end{equation}
%
\begin{figure*}[t!]
\centering
 \subfloat[][robot]{
   \includegraphics[height=35mm]{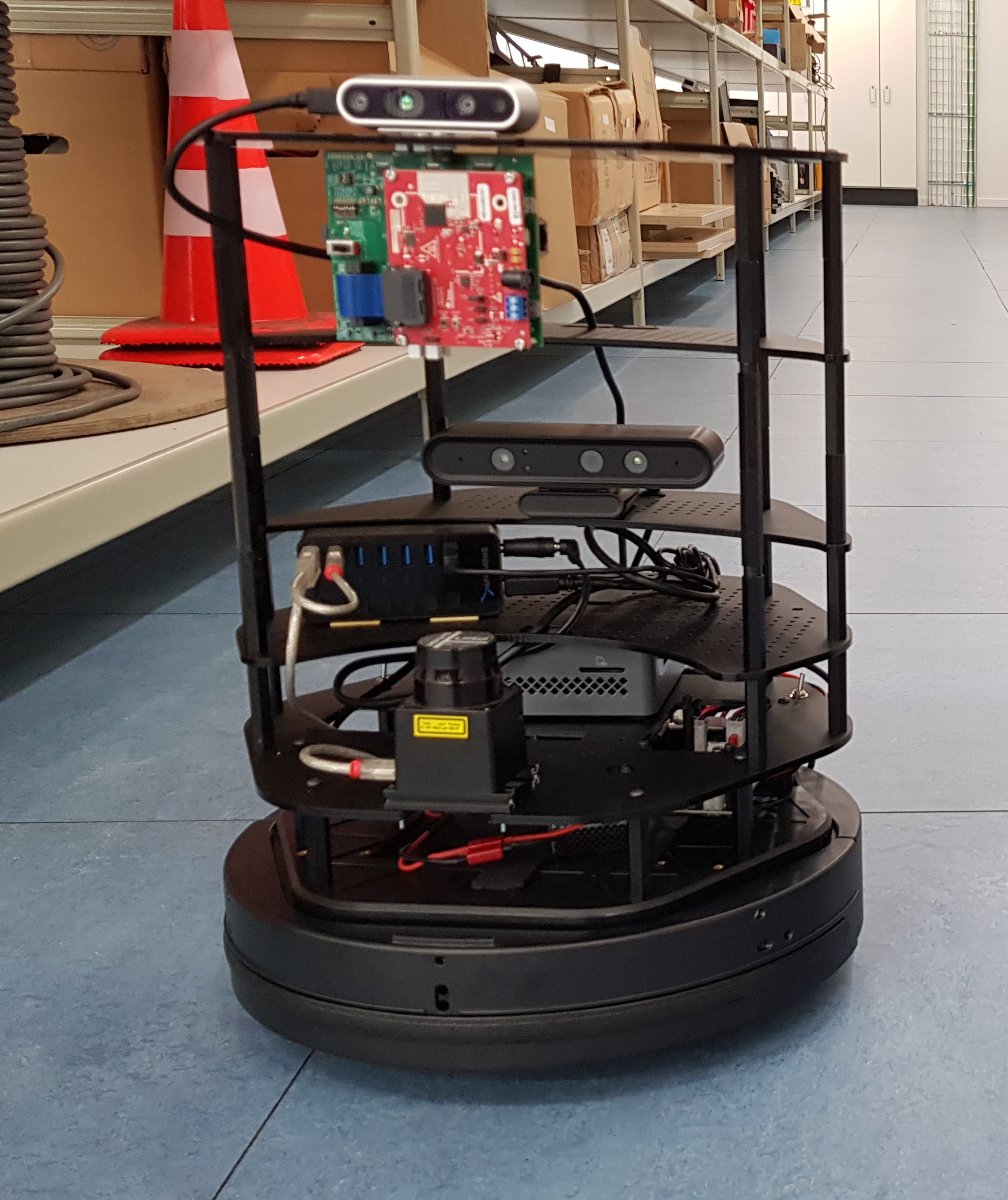}
   \label{fig:turtle}
 }
 \subfloat[][camera]{
   \includegraphics[height=35mm]{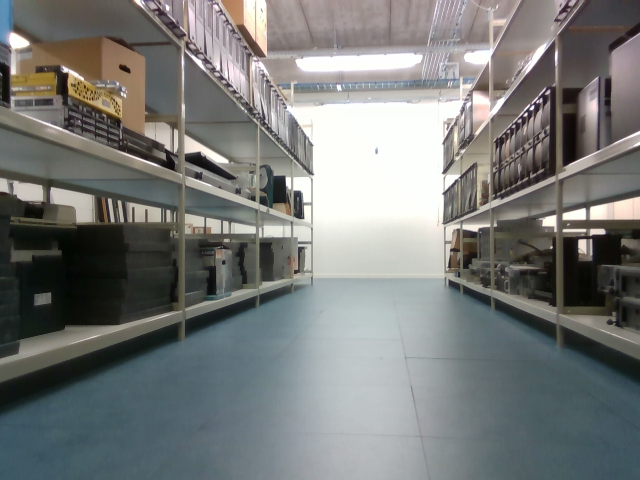}
 }
 \subfloat[][range-doppler]{
   \includegraphics[height=40mm]{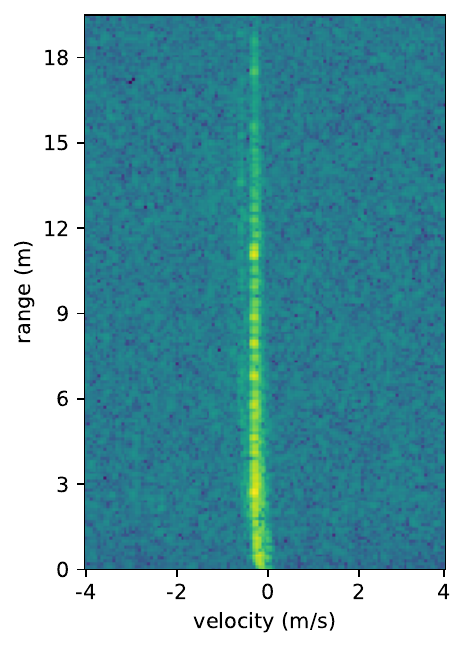}
 }
 \subfloat[][lidar]{
   \includegraphics[height=40mm]{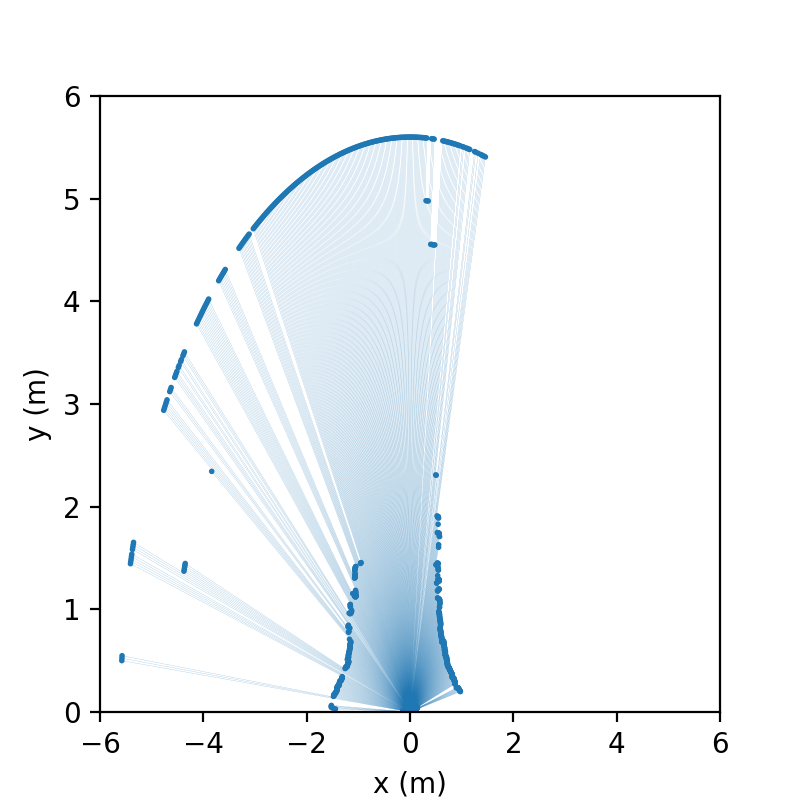}
 }
 \caption{Our dataset is captured with a Turtlebot 2i (a) equipped with an Intel Realsense D435, 79 GHz TI mmWave radar and Hokuyo URG lidar. For our experiments we use the RGB camera images (b), range-doppler maps from the radar (c) and lidar scans (d).}
\end{figure*}
From this we are interested in the posterior over the belief state, i.e. the latent code, $P(\tilde{\boldsymbol{s}} | \tilde{\boldsymbol{o}}, \tilde{\boldsymbol{a}})$. Note that we use tildes to designate sequences of the corresponding variable and that the belief state does not need to correspond to some physical state description. This state space can be purely abstract and does not necessarily have any meaning outside the model. As in many Bayesian filters, deriving the posterior distribution in general is not tractable under a normal Bayesian inference scheme, so we will apply variational inference to estimate the posterior distribution through variational autoencoding. In contrast to Kalman filtering, we do not need to explicitly state the motion or observation models. Instead we define three neural networks: a prior network, a posterior network and a likelihood network. Where each network estimates $P(\boldsymbol{s}_t | \boldsymbol{s}_{t-1}, \boldsymbol{a}_{t-1})$, $Q(\boldsymbol{s}_{t} | \boldsymbol{s}_{t-1}, \boldsymbol{a}_{t-1}, \boldsymbol{o}_t)$ and $P(\boldsymbol{o}_t | \boldsymbol{s}_t)$, representing the Markov chain model from Figure~\ref{fig:generative model}, the interplay between the prior and posterior models allow for the learning of the system dynamics. We use Equation~\ref{eq:joint} in order to derive the distributions to an estimate for every time step. The distribution $Q$ emphasizes the fact that the posterior distribution is a variational distribution, indicating that it is only an approximation of the true posterior. These networks cooperate as described in Figure~\ref{fig:model}. The prior model takes a sample of the previous state estimate and the previously taken action and uses this to predict a prior distribution over the latent state samples for the current timestep. Similarly the posterior model takes a sample of the previous state distribution, the previous action and the current observation to generate the next state distribution. We parameterize the state distributions as multivariate Gaussian distributions with diagonal covariance matrices, allowing for efficient sampling through the reparametrization trick~\cite{Kingma2014}. The likelihood model takes a state sample and decodes it back to an observation. Conforming to the free energy principle we optimize these models by minimizing their variational free energy~\cite{Friston2013life}, i.e. we minimize the Kulback-Leibler (KL) divergence between prior and posterior while also minimizing the negative log likelihood of the observations:

\begin{equation}
\label{eq:free energy}
\begin{split}
    \mathcal{L} &= \sum_{t}D_{KL} \big{[} Q( \boldsymbol{s}_t | \boldsymbol{s}_{t-1}, \boldsymbol{a}_{t-1}, \boldsymbol{o}_t) ||  P( \boldsymbol{s}_t | \boldsymbol{s}_{t-1}, \boldsymbol{a}_{t-1}) \big{]}\\
    &\qquad - \log P (\boldsymbol{o}_t | \boldsymbol{s}_t)
\end{split}
\end{equation}
This can be interpreted as a reconstruction loss, which forces the latent state to contain all information of the sensory observations, while the KL term acts as a complexity term, forcing good representations for predicting dynamics. For a more in depth discussion of the application of this type of deep neural networks to the free energy formalism we refer the reader to \cite{catal2020frontiers}. This is also similar to a Variational Auto-Encoder (VAE), but over time sequences of data and with a learnt prior instead of a standard normal prior distribution~\cite{Kingma2014}. 
\newline \newline
For LatentSLAM, we use the posterior model to infer the belief state distribution for each observation. Then, we use the mean of this distribution as template for a local view cell. This allows us to rely on dynamics aware features to perform the template matching. We calculate the similarity score $\delta(t_1,t_2)$ comparing this latent template $t_1$ to another local view cell $t_2$ as:

\begin{equation}
\label{eq:similarity}
    \delta(t_1, t_2) = 1 - \frac{t_1 \cdot t_2}{||t_1|| \cdot ||t_2||}
\end{equation}
This boils down to 1 minus the cosine similarity. The lower the score, the higher the similarity, with a score of zero when $t_1 = t_2$. In case we have multiple sensor modalities, we train a posterior model for each sensor. We can then fuse multiple sensors by concatenating the latent templates, or by calculating the score for each modality, and taking the maximum score.

\section{Dataset}
\label{sec:dataset}

For the experiments we collect a dataset with a Turtlebot 2i equipped with an Intel Realsense D435 camera, a TI IWR1443 mmWave radar and a Hokuyo URG lidar (Fig.~\ref{fig:turtle}). We teleoperate the robot throughout our industrial IoT lab environment, which consists of a 10m x 30m space divided into similar looking corridors by shelves.
\newline \noindent
We record the following scenarios:
\begin{enumerate}
    \item Small loops around each one of the shelve array.
    \item Big loops, close to the outer walls.
    \item 8-shaped trajectories through the aisles.
    \item Driving back and forth in a single aisle, performing U-turns.
    \item Two large trajectories, also crossing the open space.
    \item 30 minute trajectory between the aisles.
\end{enumerate}

\begin{figure}[t]
	\centering
\includegraphics[width=0.42\textwidth]{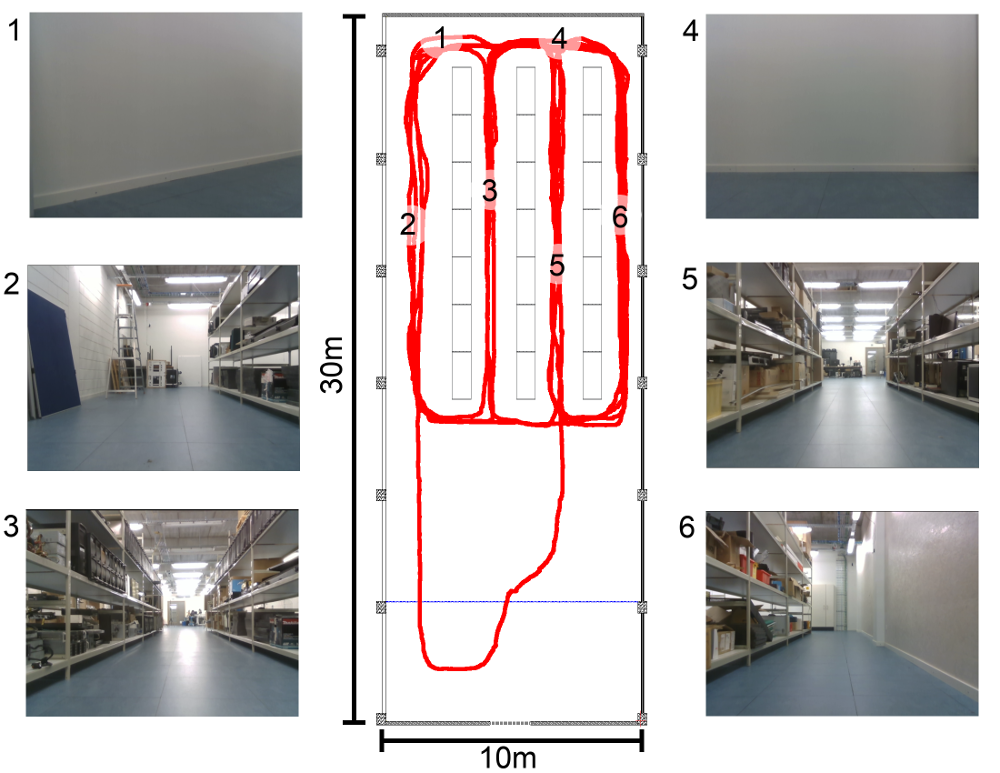}
    \caption{Map of our lab environment, together with a number of views at various location on the tracks. The robot location is often difficult to disambiguate based on camera input only (i.e. 1 vs 4 and 3 vs 5).}
    \label{fig:lab}
\end{figure}

Figure~\ref{fig:lab} shows a map of the Industrial IoT lab of IDLab Ghent, annotated with the trajectories recorded by the robot. We also add the camera view at various locations on the map. It is clear that the two middle aisles (e.g. at location 3 and 5 in Fig.~\ref{fig:lab}) are difficult to disambiguate from the camera data, as well as the aisle ends where the robot is facing a white wall.

We record all data at 10Hz, consisting of 640x480 RGB and depth images from the camera, range-doppler and range-azimuth maps from the radar, lidar scans, odometry and velocity twist commands. In total we collected almost two hours of data totaling 65Gb.

\begin{figure*}[t]
\centering
 \subfloat[][camera]{
   \includegraphics[height=35mm]{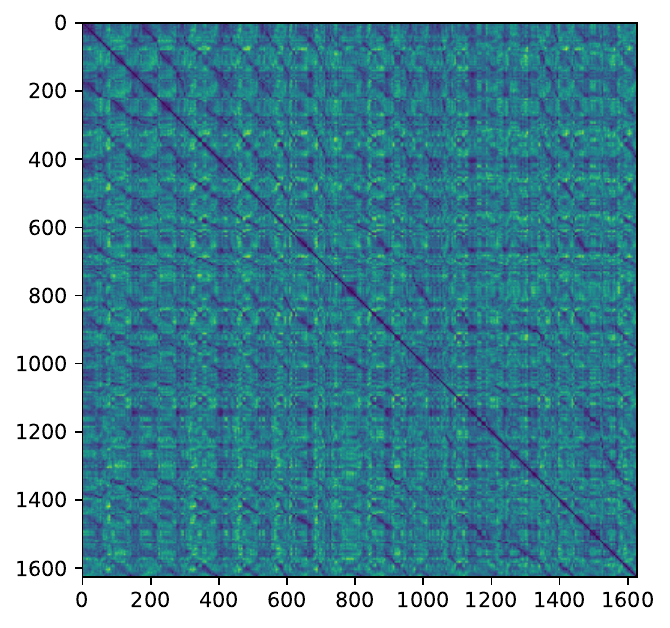}
 }
 \subfloat[][range-doppler]{
   \includegraphics[height=35mm]{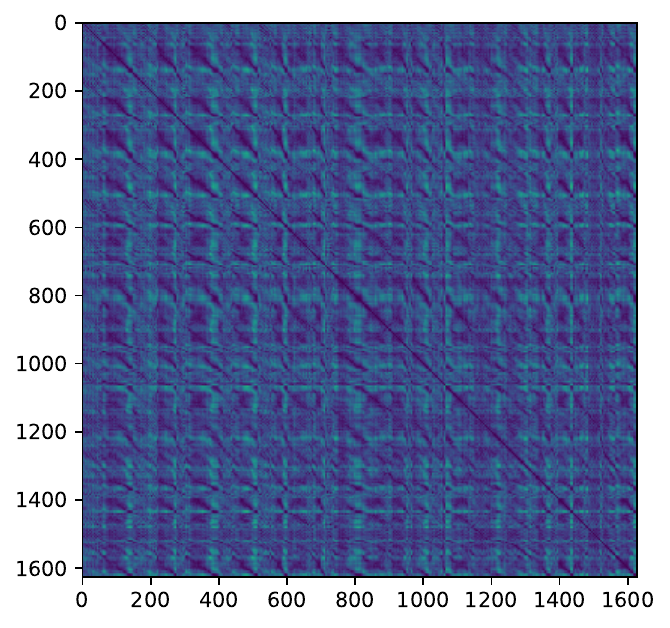}
 }
 \subfloat[][lidar]{
   \includegraphics[height=35mm]{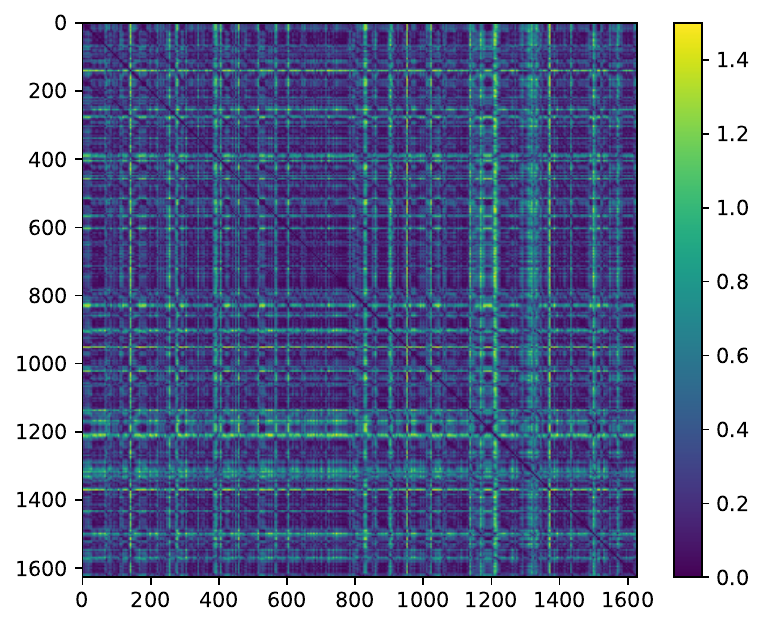}
 }
 \caption{We plot the scores $\delta(t_1,t_2)$ for the LatentSLAM templates of all view cells of the scenario 6 map, which illustrates the confusion between view cells for different sensor modalities. Different sensors provide complimentary information, for example lidar performs bad inside the aisles, but best at the aisle ends.}
 \label{fig:confusing}
\end{figure*}

\section{Experiments}
\label{sec:experiments}

\subsection{Model training}

We instantiate our neural network models depicted in Figure~\ref{fig:model} as follows. As prior model, we use an LSTM~\cite{schmidhuber1997} with 128 hidden units to process concatenated previous state and action vectors. Camera images are processed at a 160x120 resolution by a 6-layer convolutional pipeline with 32, 32, 64, 64, 128 and 128 filters of 3x3 which reduces the spatial resolutions with a factor 2 when the number of filters increases. The resulting features are flattened and concatenated with the previous state and action and processed by another fully connected layer with 128 hidden units. The likelihood model is a mirrored convolutional pipeline that upsamples by nearest neighbour interpolation to form the reconstructed observation. The posterior and likelihood models for range-doppler data are similar, but now operating on 192x128 range-doppler maps with 8, 16, 32, 64 and 128 convolutional filters. The lidar scans are 1 dimensional vectors of size 726, for which we use a fully connected neural network with two layers of 512 hidden units for both the posterior and likelihood model. As actions we use raw twist commands, containing the linear and angular velocities. Our state space is parameterized as a 32 dimensional multivariate Gaussian distribution, hence state samples are vectors of size 32 and distributions are represented by vectors of size 64, containing the 32 means and 32 standard deviations of the Gaussian distributions. To force the standard deviations to be positive, we use a SoftPlus~\cite{dugas2000}. For all layers we use a leaky ReLU activation non-linearity~\cite{Maas2013RectifierNI}.

We train our models on a subset of the dataset, i.e. scenarios 1, 2 and 4. During training, we sample mini-batches containing 32 random subsequences of 10 time steps. We train for 500 epochs using the Adam optimizer~\cite{Kingma14adam} with initial learning rate 1e-4. To avoid posterior collapse in the range-doppler maps we use Geco~\cite{Rezende2018GeneralizedEW} as a constraint based optimization scheme.

\subsection{LatentSLAM results}

We evaluate our LatentSLAM approach on scenarios 3, 5 and 6 of the dataset which were held-out during training. Scenario 3 contains two runs of an 8-shaped trajectory that illustrates basic map building and loop closing. Scenarios 5 and 6 consist of longer runs through the environment. In scenario 5 the robot also drives through the open space of the lab (Fig.~\ref{fig:lab}), to check the robustness against sensory inputs that are slightly different from the train-distribution.

We compare against RTAB-Map (Real-Time Appearance-Based Mapping)~\cite{Labbe2019}, an RGBD graph-based SLAM approach, and the original RatSLAM~\cite{Milford2004} approach in Table~\ref{tab:results}. In order for RTAB-Map to work, we had to map the different aisles separately in 3 runs, and then combine the maps offline. Otherwise, the graph optimization tended to fail due to incorrect matches, mapping different aisles onto each other. Since we only mapped the aisles, the loop in the open space is also missing in scenario 5 for RTAB-Map.

Since in LatentSLAM, we use an odometry signal based on the velocity commands used to control the robot. This mimics proprioceptive feedback through efferent copies in the brain~\cite{Jeannerod2003}, and makes our approach independent from any wheel encoders or IMU sensors. The resulting raw odometry trajectories are shown in Table~\ref{tab:results}, row 3.

For RatSLAM and LatentSLAM, we tune the matching threshold $\delta_{match}$ to a value of 0.05.
When using the camera sensor (Table~\ref{tab:results}, row 4), LatentSLAM performs on par with RatSLAM for scenario 3, however, especially on the longer scenarios 5 and 6 LatentSLAM collapses the maps less due to incorrect template matches. In scenario 6 LatentSLAM has problems differentiating between locations that look similar for the camera. Also note that, although not part of the training data, LatentSLAM is able to cover the trajectory through the open space in scenario 5, and to correctly close the loop. Table~\ref{tab:results}, row 5 shows that LatentSLAM also works when using latent states trained on range-doppler maps. However, this configuration has even more difficulties to disambiguate between the different aisles on the longer sequences. 

However, when we fuse both camera and range-doppler by concatenation of their latent vectors, LatentSLAM performs significantly better on scenario 5 and 6 (Table~\ref{tab:results}, row 6). Now the different aisles are clearly distinguished, and only at the aisle end points we observe confusion. This can be explained intuitively, as at the end of the two middle aisles the robot will be facing a white wall at the same range. When also combining a latent code trained on lidar scans, the performance improves slightly (Table~\ref{tab:results}, row 7). Here we found it was best to fuse using the highest scoring modality, either radar and camera or lidar, as the lidar scans are very similar inside the aisles, which leads to more confusion when concatenating. Also note that LatentSLAM maps, including stored templates, on scenario 6 result in a size of about 2MB, whereas the metric map from RTAB-Map takes over 850MB.

Figure~\ref{fig:confusing} shows the confusion between different view cells of different sensor modalities. From this it is clear that overall the camera is best for differentiating the view cells, but range-doppler and lidar do provide some extra, complimentary information. For example lidar provides higher scores to disambiguate robot poses at the end of the aisles.
 
\subsection{Ablation study}

Deep neural networks are often claimed to be powerful image feature detectors, even when trained on other tasks such as image classification on ImageNet. To test this, we also attempted using feature embeddings from a pre-trained ResNet-18 neural network ~\cite{He2016} for LatentSLAM. We found that when using these embeddings, the system was much more sensitive to the value of the matching threshold $\delta_{match}$, which required to be at a much lower value in order to successfully detect loop closures. This results in many more view cells in the map, and quick collapsing when the threshold is put slightly higher. Also, these pre-trained embeddings have many more dimensions than our LatentSLAM models (1024 vs 32).

\begin{figure}[b]
\centering
 \subfloat[][]{
   \includegraphics[width=35mm]{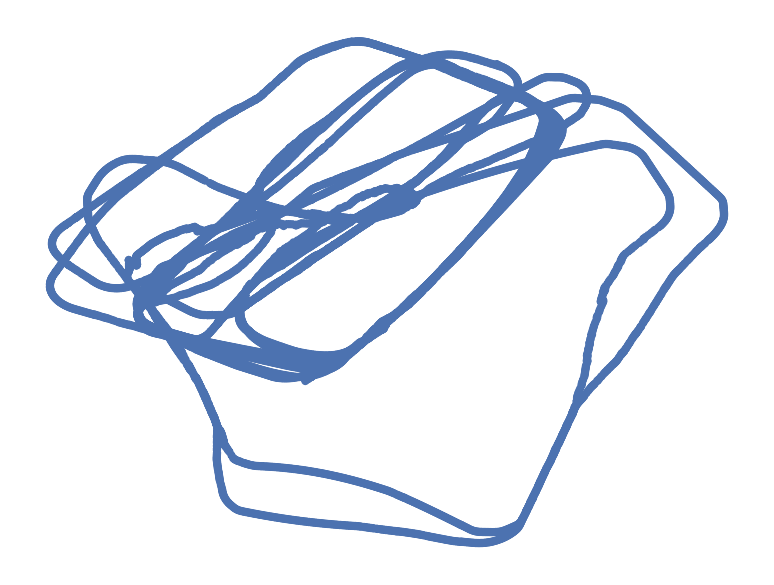}
 }
 \subfloat[][]{
   \includegraphics[width=35mm]{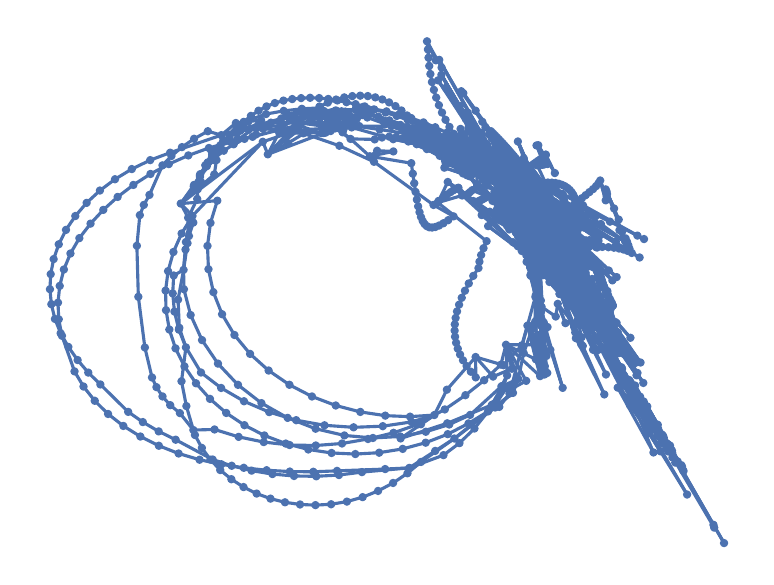}
 }
 \caption{Ablation with ResNet-18 features, which is sensitive to the matching threshold $\delta_{match}$ 0.003 (a) vs 0.004 (b)}
\end{figure}

Since our posterior models output multivariate Gaussian distributions, we also investigated scoring templates using the KL divergence, as opposed to cosine similarity between their means. This also works, but also here tuning the threshold turned out to be more difficult, as the KL divergence ranges from zero to infinity.

\section{Conclusion and Future Work}
\label{sec:conclusion}

In this paper, we proposed LatentSLAM, a biologically inspired SLAM algorithm that adapts RatSLAM~\cite{Milford2004} to use learnt latent state descriptors. We propose a predictive generative model that is able to learn compact representations of different sensor modalities, which we applied on camera, range-doppler and lidar data. We showed that LatentSLAM yields better maps in a challenging environment, and that multiple sensors can be combined in latent space to increase robustness.

We believe it opens some interesting avenues for future work. First, we want to investigate better fusion algorithm besides manually concatenating latent states or taking the maximum score. For example, we might use a weighted scoring algorithm similar to~\cite{Jacobson2012}. An other idea is to incorporate the sensor fusion during model training, by sharing a single prior model and fusing the different observations in a single posterior model that learns a shared state space.
Second, we currently only use the posterior model for localization and mapping. However, in previous work the prior model has also proved to be useful, for example for anomaly detection~\cite{catal2020iros} or control~\cite{catal2020icassp}. In our case the uncertainty of the prior model could for example be used for detecting changes in the environment, or for driving exploration.
Finally, our current approach requires training upfront on sequences of data recorded in the environment. In future work we will investigate whether we can train the models online while exploring and mapping an area. 

\begin{table*}[t!]
  \caption{Resulting maps created by LatentSLAM using camera, range-doppler, lidar or a combination thereof.  We compare against RTAB-Map~\cite{Labbe2019}, RatSLAM~\cite{Milford2004} and raw odometry.}
  \centering
  \begin{tabular}{*{5}{c}}
    & scenario 3A & scenario 3B & scenario 5 & scenario 6\\
    \hline
    \rotatebox{90}{\hspace{0.4cm}RTAB-Map~\cite{Labbe2019}}&
    \includegraphics[width=35mm]{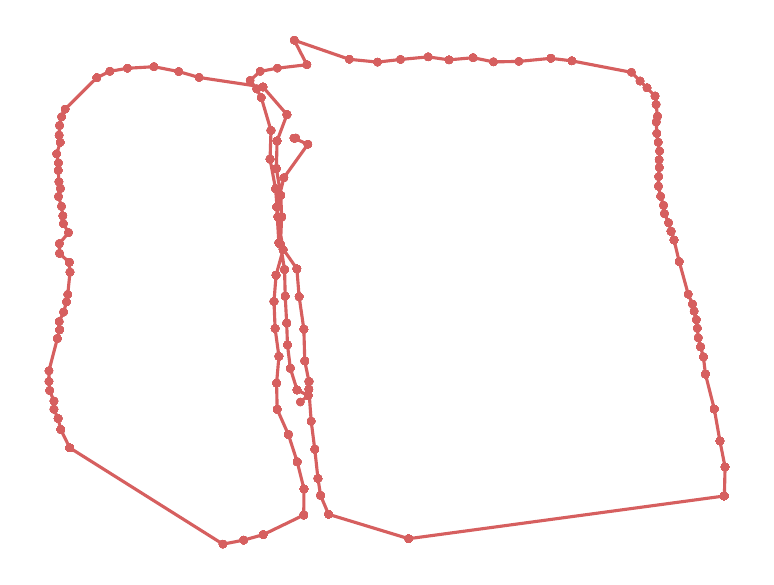} &
    \includegraphics[width=35mm]{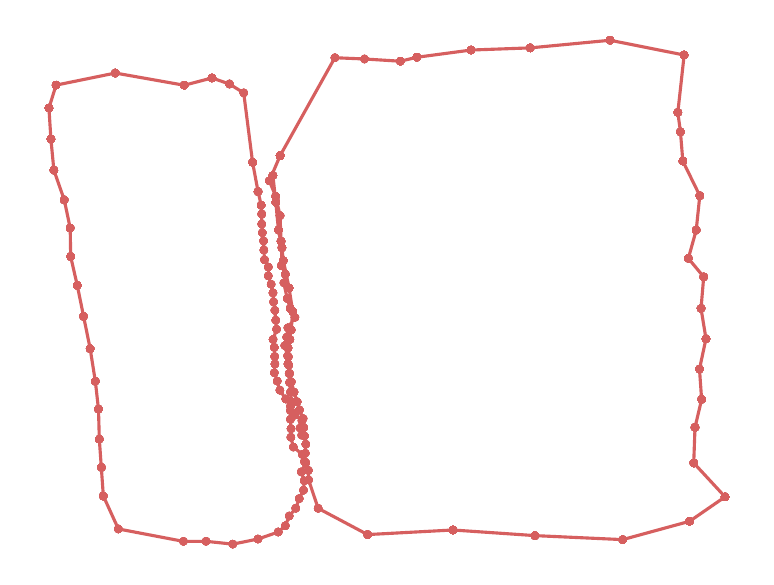} &
    \includegraphics[width=35mm]{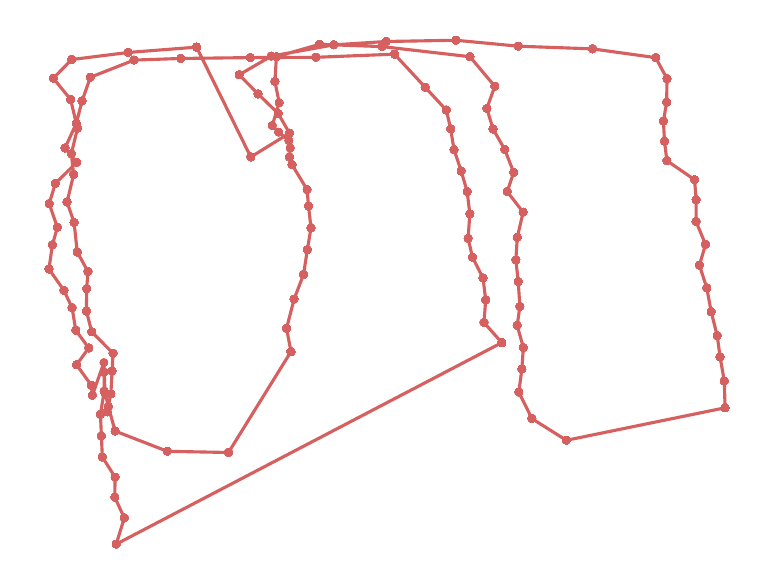} &
    \includegraphics[width=35mm]{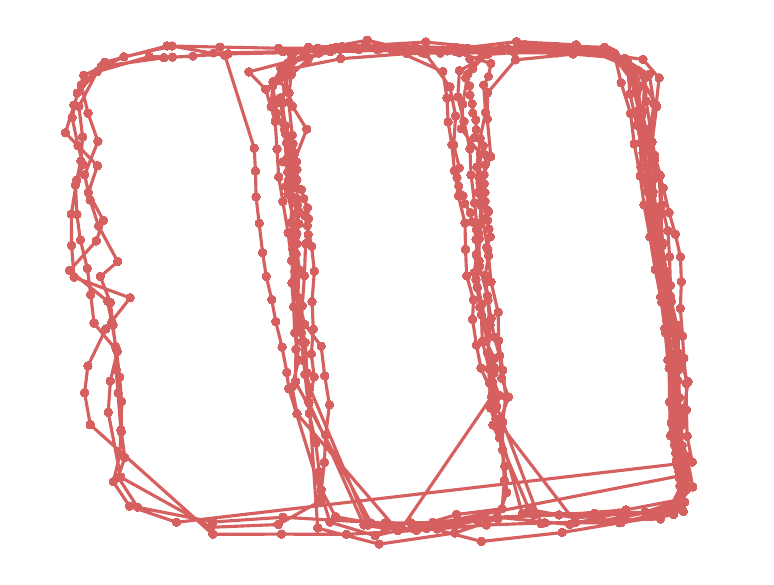}\\
    \hline
    \rotatebox{90}{\hspace{0.4cm}RatSLAM~\cite{Milford2004}}&
    \includegraphics[width=35mm]{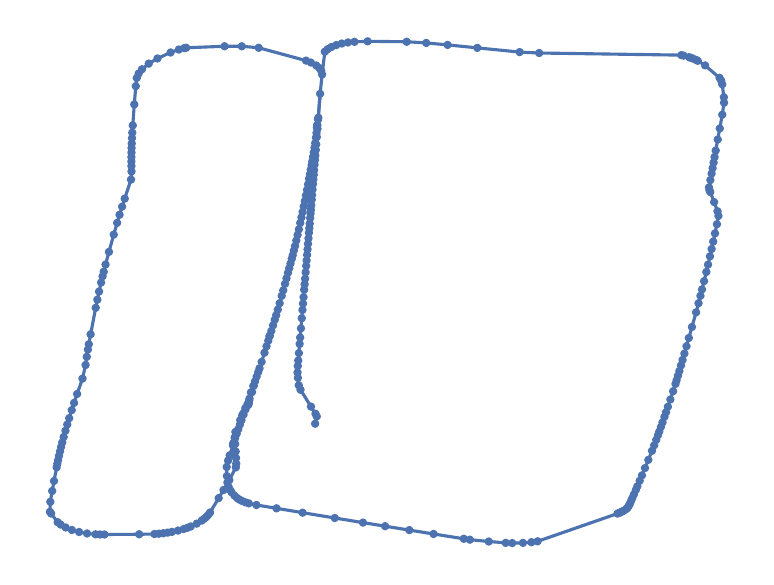} &
    \includegraphics[width=35mm]{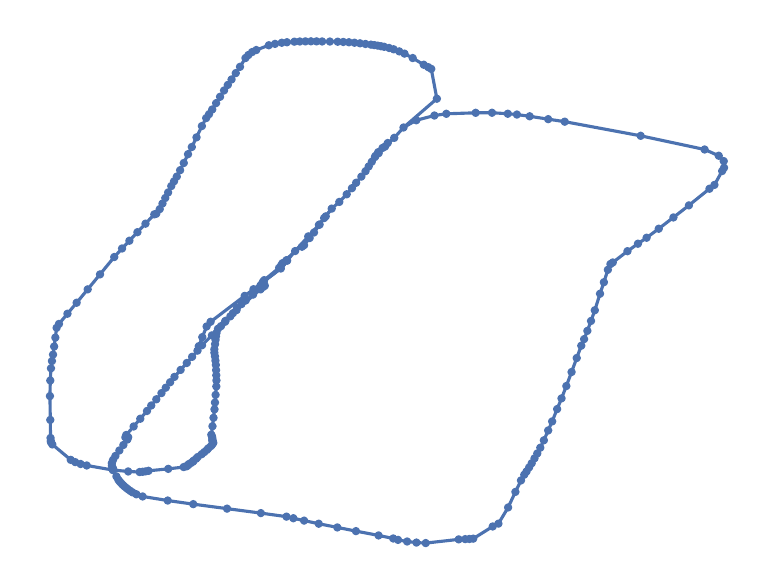} &
    \includegraphics[width=35mm]{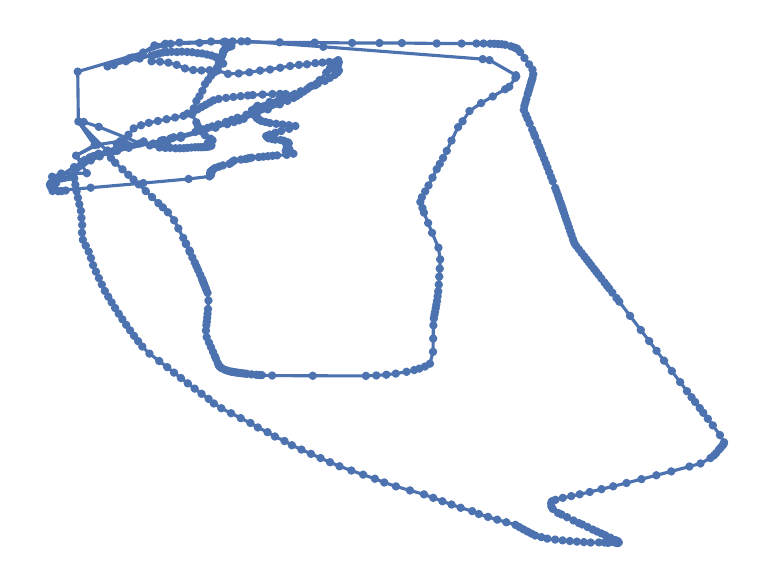} &
    \includegraphics[width=35mm]{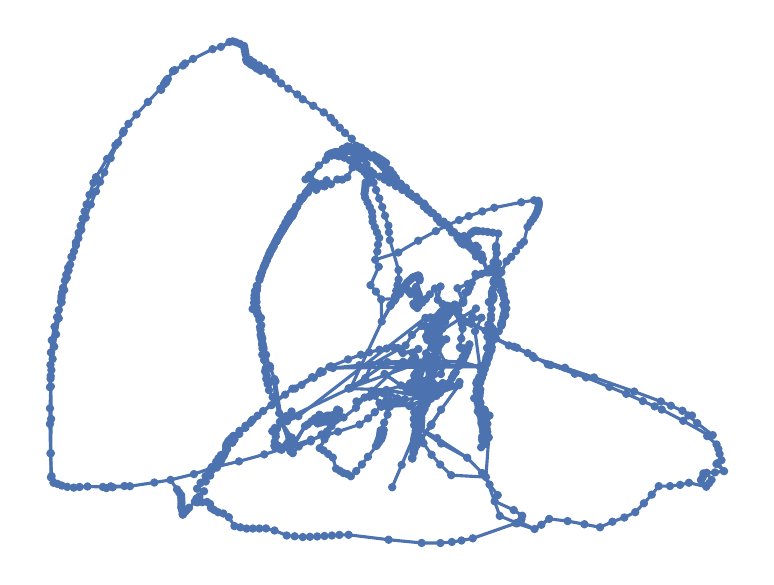}\\ 
    \hline
    \rotatebox{90}{\hspace{0.7cm}odometry}&
    \includegraphics[width=35mm]{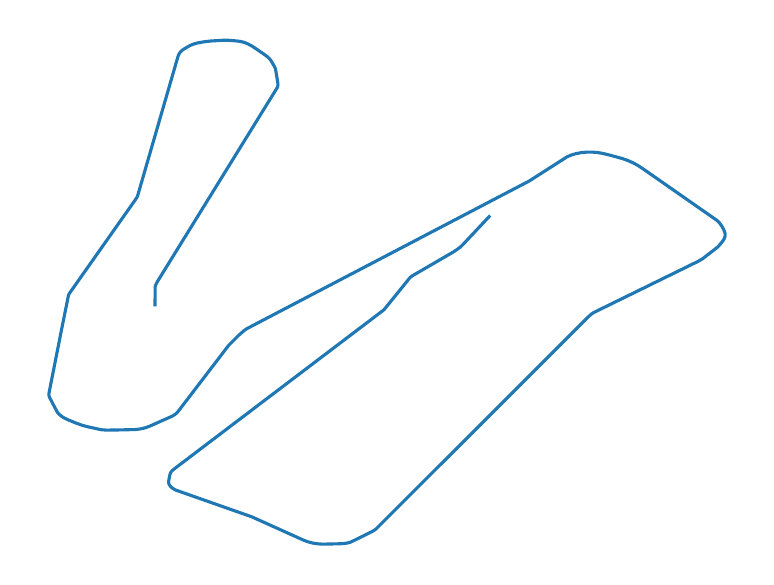} &
    \includegraphics[width=35mm]{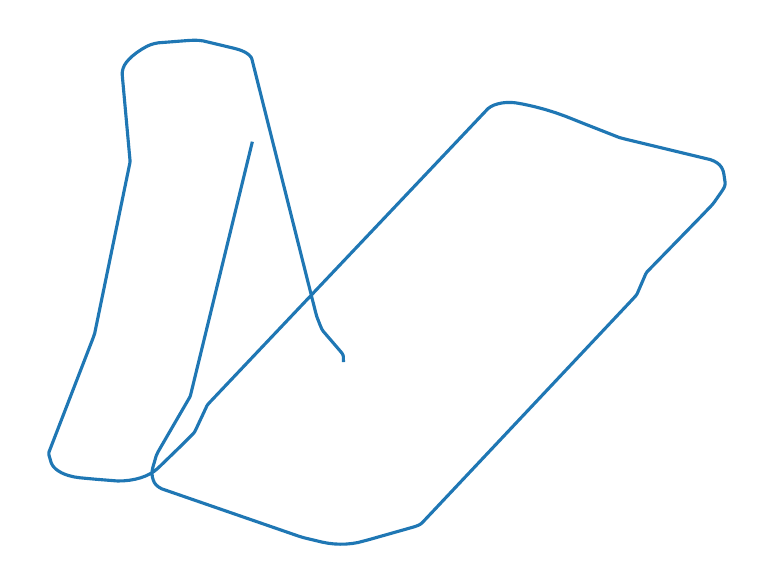} &
    \includegraphics[width=35mm]{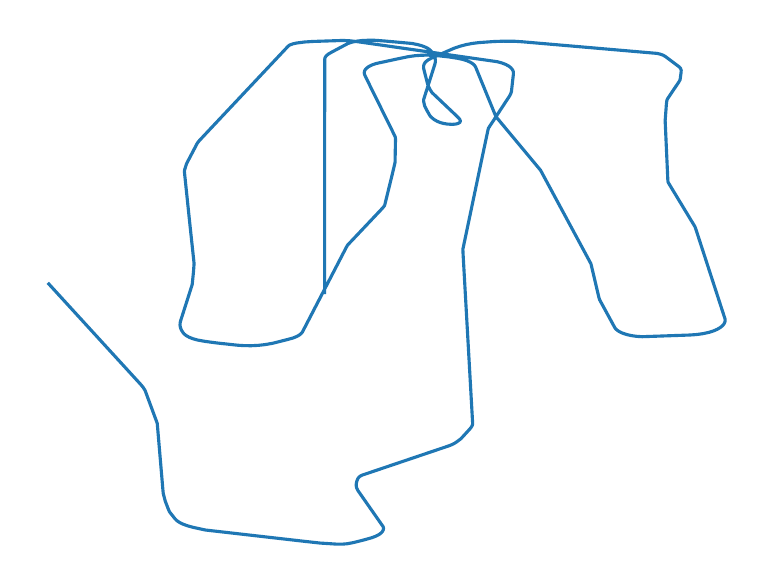} &
    \includegraphics[width=35mm]{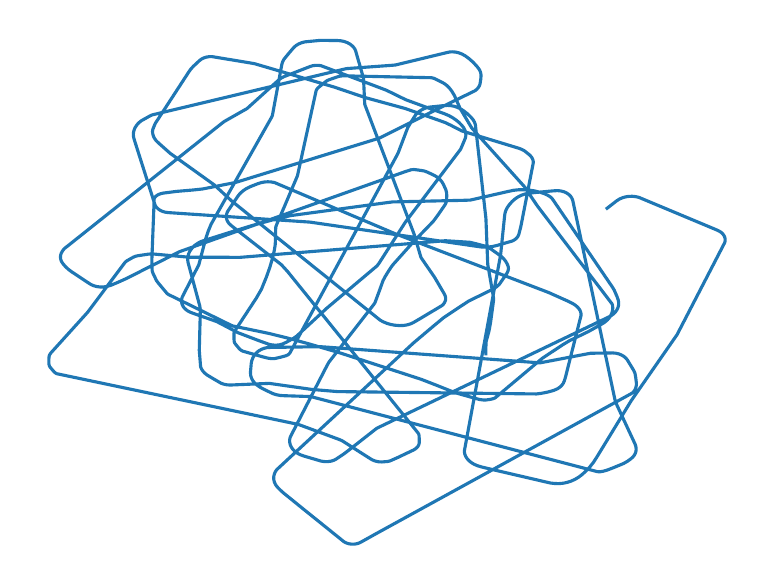}\\
    \hline
    \rotatebox{90}{\parbox{3cm}{\centering LatentSLAM \\ camera}}&
    \includegraphics[width=35mm]{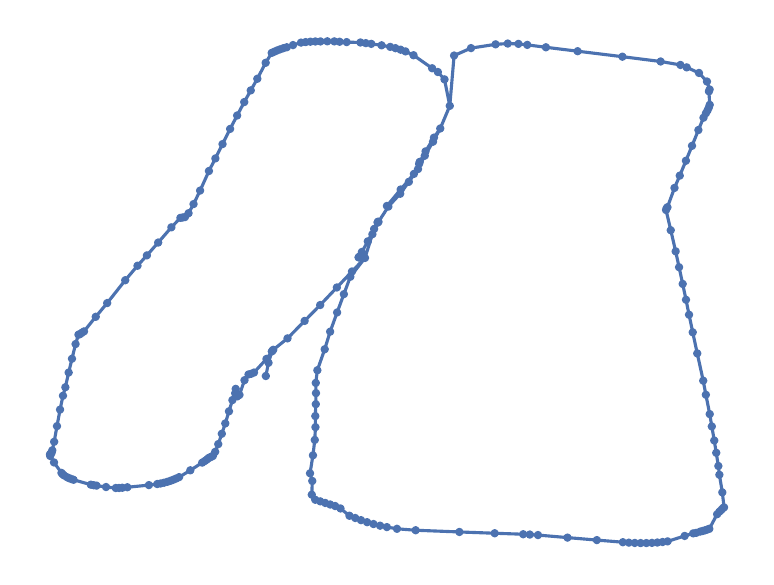} &
    \includegraphics[width=35mm]{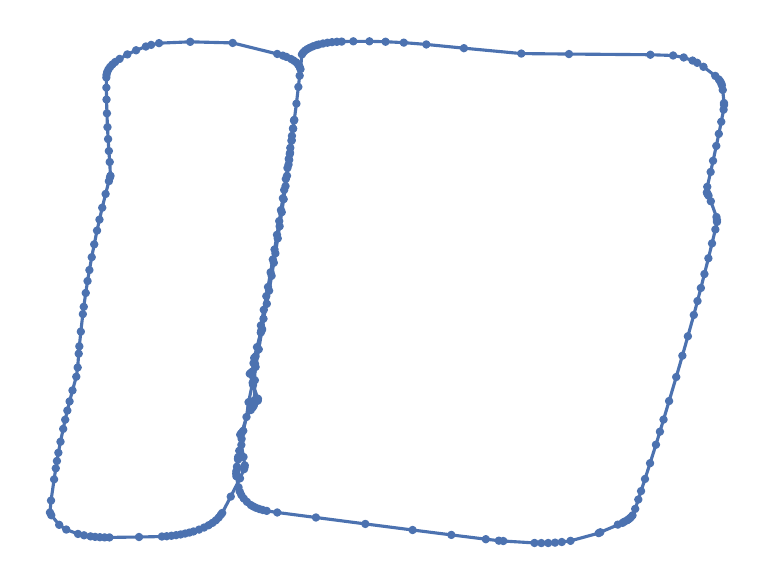} &
    \includegraphics[width=35mm]{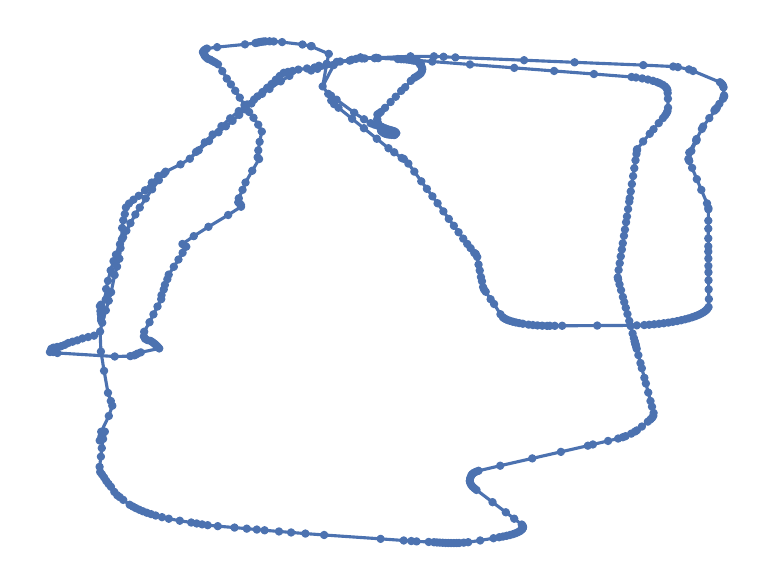} &
    \includegraphics[width=35mm]{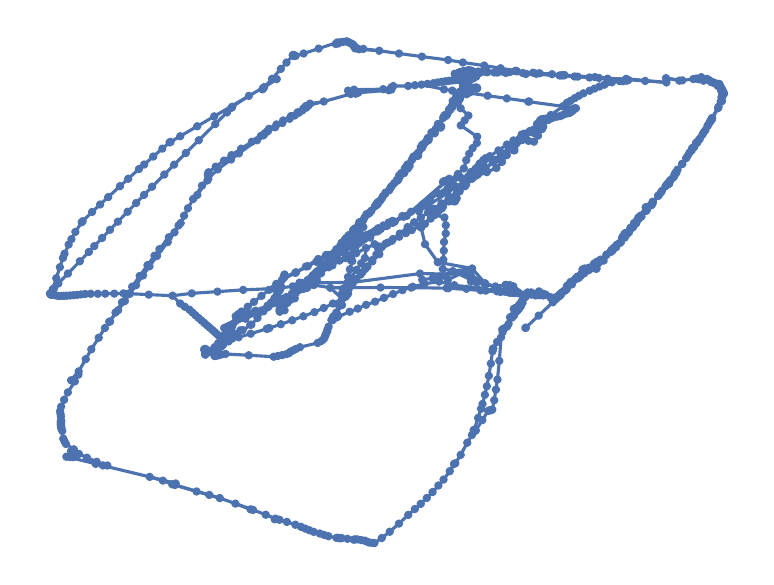}\\
    \hline
    \rotatebox{90}{\parbox{3cm}{\centering LatentSLAM \\ range-doppler}}&
    \includegraphics[width=35mm]{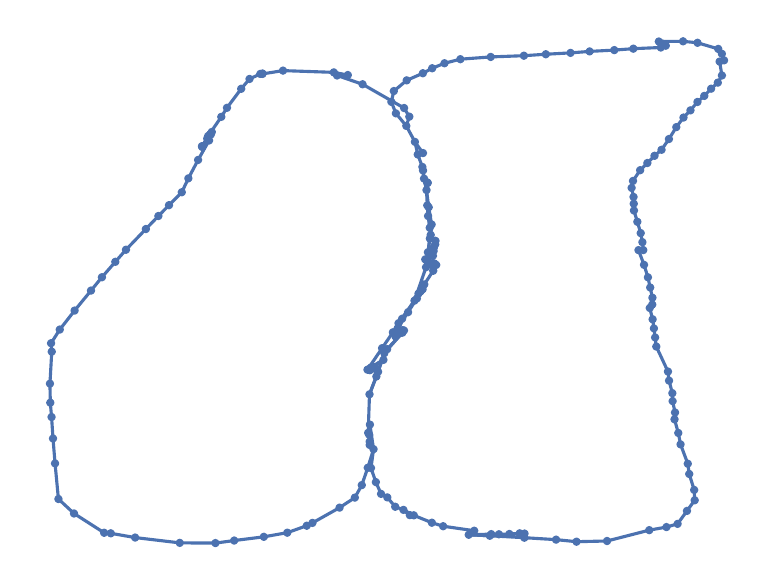} &
    \includegraphics[width=35mm]{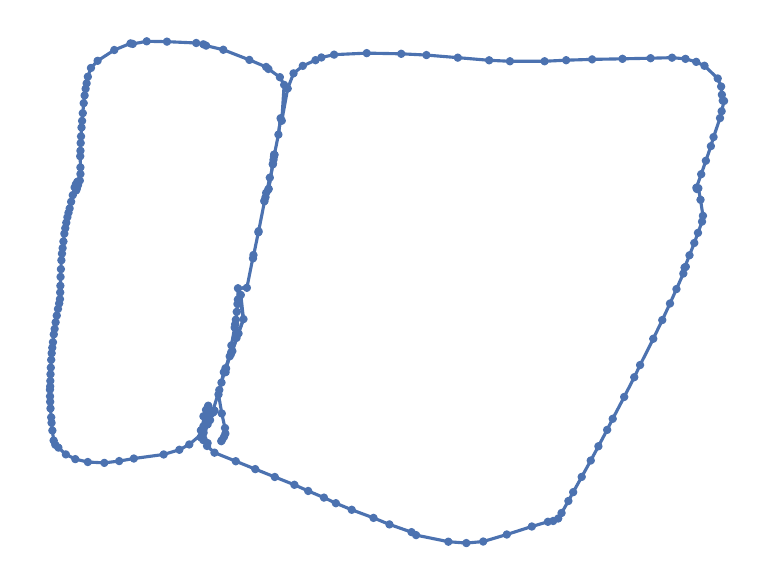} &
    \includegraphics[width=35mm]{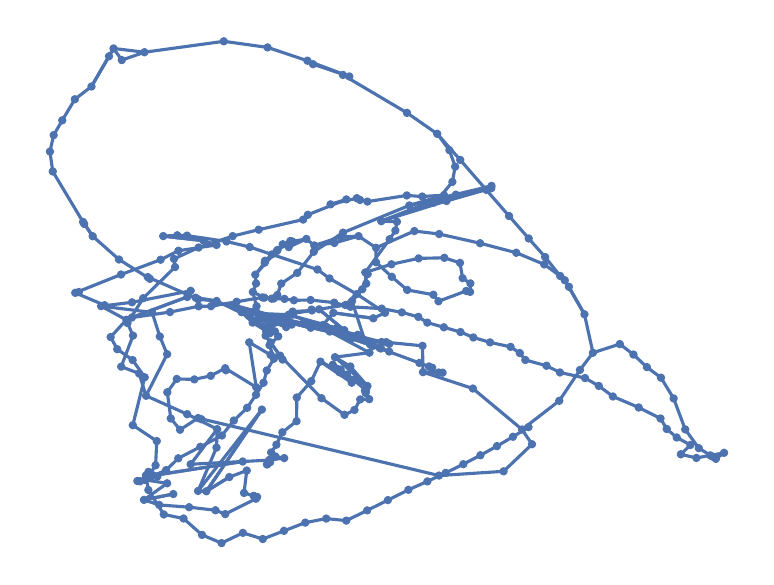} &
    \includegraphics[width=35mm]{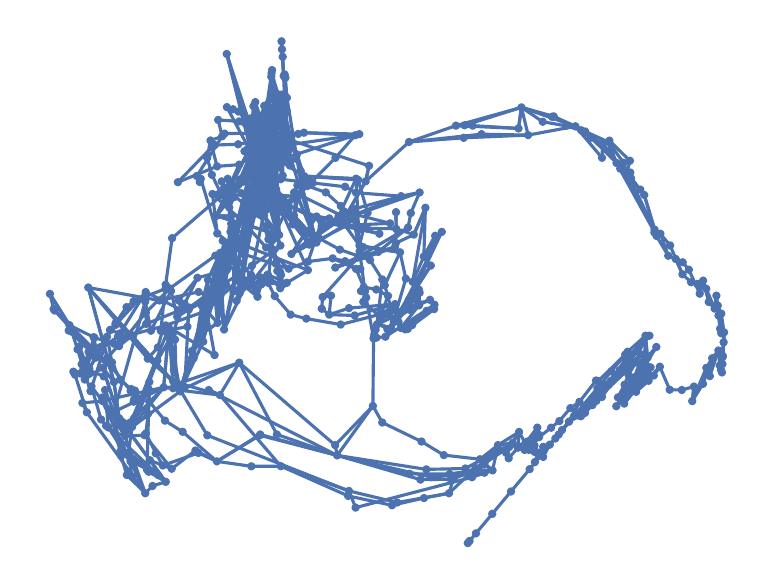}\\
    \hline
    \rotatebox{90}{\parbox{3cm}{\centering LatentSLAM \\ camera $\oplus$ range-doppler}}&
    \includegraphics[width=35mm]{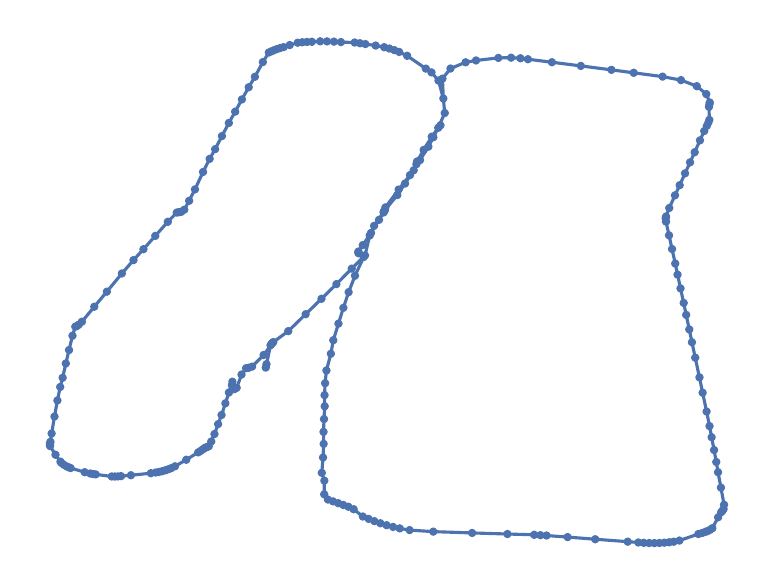} &
    \includegraphics[width=35mm]{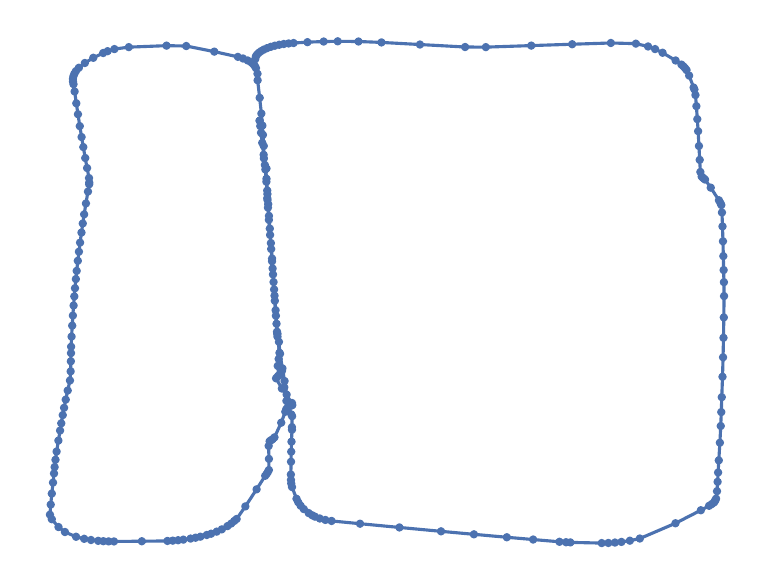} &
    \includegraphics[width=35mm]{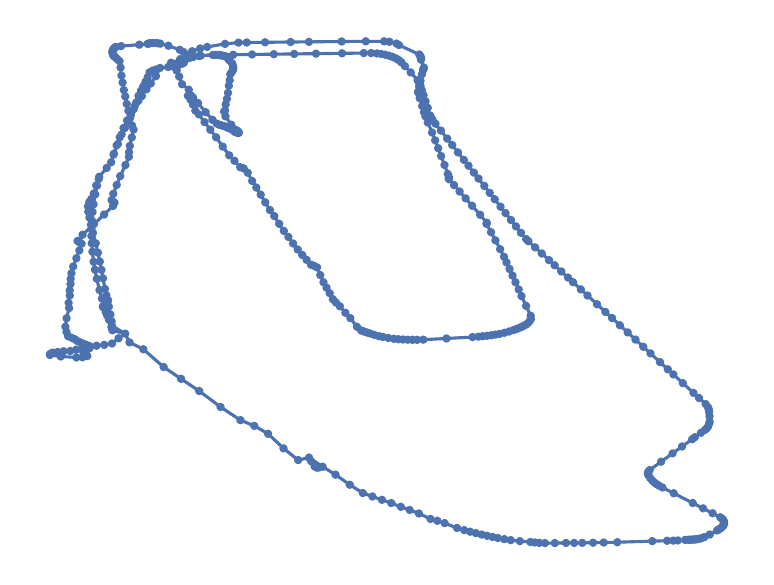} &
    \includegraphics[width=35mm]{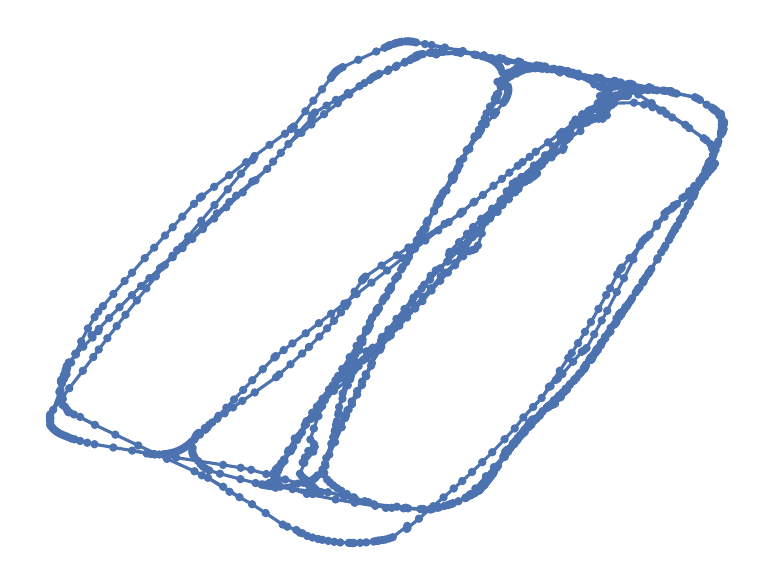}\\
    \hline
    \rotatebox{90}{\parbox{3cm}{\centering LatentSLAM \\ max(camera $\oplus$ ran-dop), lidar)}}&
    \includegraphics[width=35mm]{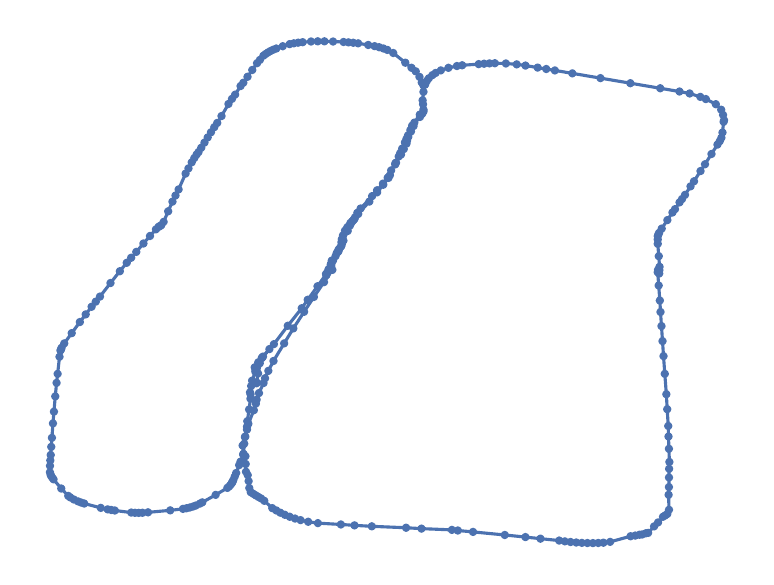} &
    \includegraphics[width=35mm]{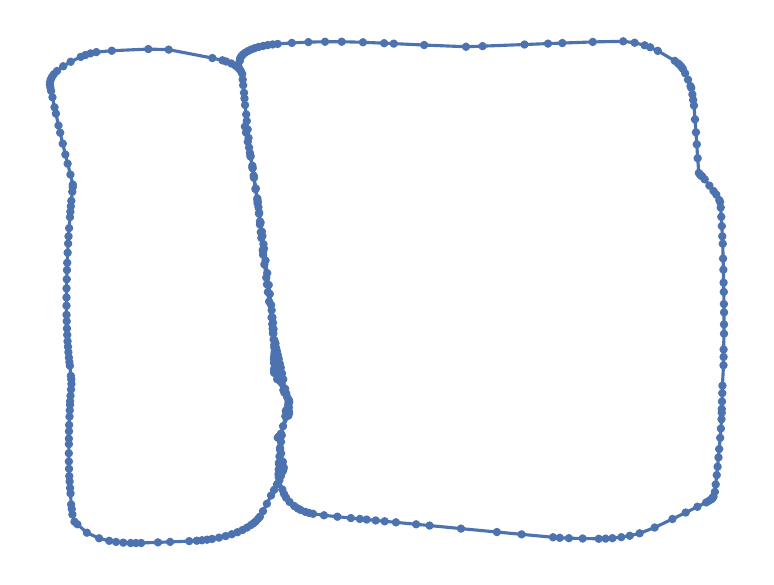} &
    \includegraphics[width=35mm]{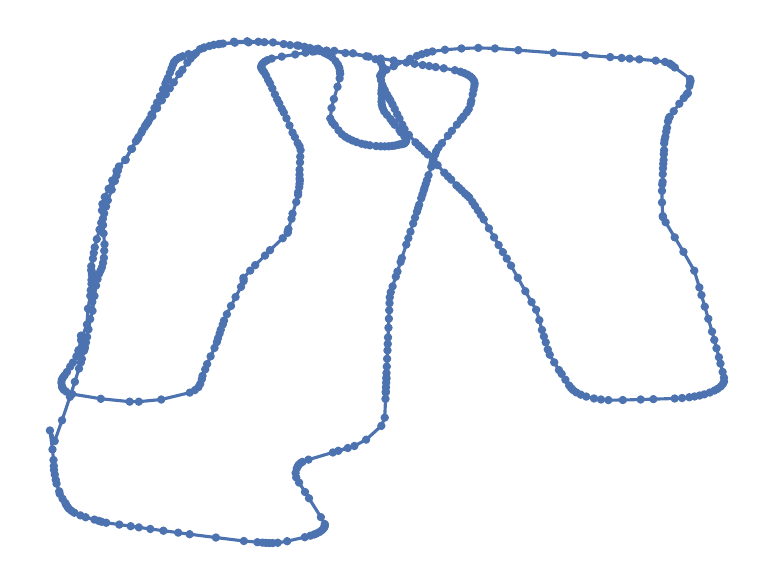} &
    \includegraphics[width=35mm]{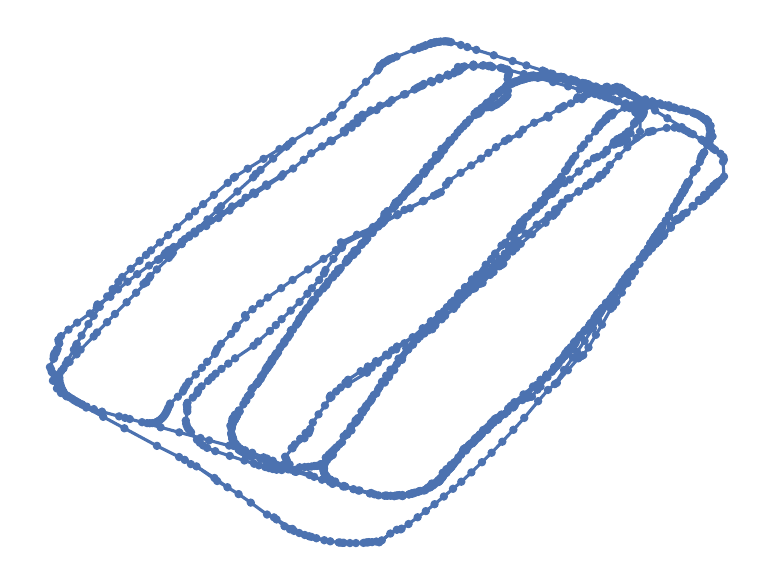}\\
    \hline
  \end{tabular}
  \label{tab:results}
\end{table*}

\section*{Acknowledgments}

O.\c{C}. is funded by a Ph.D. grant of the Flanders Research  Foundation (FWO). This research received funding from the Flemish Government under the ``Onderzoeksprogramma~Artificiële~Intelligentie~(AI)~Vlaanderen'' program.
\cleardoublepage

\bibliographystyle{IEEEtran}
\bibliography{references}

\begin{thebibliography}{10}
\providecommand{\url}[1]{#1}
\csname url@rmstyle\endcsname
\providecommand{\newblock}{\relax}
\providecommand{\bibinfo}[2]{#2}
\providecommand\BIBentrySTDinterwordspacing{\spaceskip=0pt\relax}
\providecommand\BIBentryALTinterwordstretchfactor{4}
\providecommand\BIBentryALTinterwordspacing{\spaceskip=\fontdimen2\font plus
\BIBentryALTinterwordstretchfactor\fontdimen3\font minus
  \fontdimen4\font\relax}
\providecommand\BIBforeignlanguage[2]{{%
\expandafter\ifx\csname l@#1\endcsname\relax
\typeout{** WARNING: IEEEtran.bst: No hyphenation pattern has been}%
\typeout{** loaded for the language `#1'. Using the pattern for}%
\typeout{** the default language instead.}%
\else
\language=\csname l@#1\endcsname
\fi
#2}}

\bibitem{Cadena2016}
\BIBentryALTinterwordspacing
C.~Cadena, L.~Carlone, H.~Carrillo, Y.~Latif, D.~Scaramuzza, J.~Neira, I.~Reid,
  and J.~J. Leonard, ``Past, present, and future of simultaneous localization
  and mapping: Toward the robust-perception age,'' \emph{Trans. Rob.}, vol.~32,
  no.~6, p. 1309–1332, Dec. 2016. [Online]. Available:
  \url{https://doi.org/10.1109/TRO.2016.2624754}
\BIBentrySTDinterwordspacing

\bibitem{Artal2015}
R.~{Mur-Artal}, J.~M.~M. {Montiel}, and J.~D. {Tardós}, ``Orb-slam: A
  versatile and accurate monocular slam system,'' \emph{IEEE Transactions on
  Robotics}, vol.~31, no.~5, pp. 1147--1163, 2015.

\bibitem{Shi2020}
X.~{Shi}, D.~{Li}, P.~{Zhao}, Q.~{Tian}, Y.~{Tian}, Q.~{Long}, C.~{Zhu},
  J.~{Song}, F.~{Qiao}, L.~{Song}, Y.~{Guo}, Z.~{Wang}, Y.~{Zhang}, B.~{Qin},
  W.~{Yang}, F.~{Wang}, R.~H.~M. {Chan}, and Q.~{She}, ``Are we ready for
  service robots? the openloris-scene datasets for lifelong slam,'' in
  \emph{2020 IEEE International Conference on Robotics and Automation (ICRA)},
  2020, pp. 3139--3145.

\bibitem{Milford2004}
M.~J. {Milford}, G.~F. {Wyeth}, and D.~{Prasser}, ``Ratslam: a hippocampal
  model for simultaneous localization and mapping,'' in \emph{IEEE
  International Conference on Robotics and Automation, 2004. Proceedings. ICRA
  '04. 2004}, vol.~1, 2004, pp. 403--408 Vol.1.

\bibitem{Ball2013}
D.~Ball, S.~Heath, J.~Wiles, G.~Wyeth, P.~Corke, and M.~Milford,
  ``{OpenRatSLAM: An open source brain-based SLAM system},'' \emph{Autonomous
  Robots}, vol.~34, no.~3, pp. 149--176, 2013.

\bibitem{Gu2019}
T.~{Gu} and R.~{Yan}, ``An improved loop closure detection for ratslam,'' in
  \emph{2019 5th International Conference on Control, Automation and Robotics
  (ICCAR)}, 2019, pp. 884--888.

\bibitem{Steckel2013}
\BIBentryALTinterwordspacing
J.~Steckel and H.~Peremans, ``Batslam: Simultaneous localization and mapping
  using biomimetic sonar,'' \emph{PLOS ONE}, vol.~8, no.~1, pp. 1--11, 01 2013.
  [Online]. Available: \url{https://doi.org/10.1371/journal.pone.0054076}
\BIBentrySTDinterwordspacing

\bibitem{Fox2012}
C.~{Fox}, M.~{Evans}, M.~{Pearson}, and T.~{Prescott}, ``Tactile slam with a
  biomimetic whiskered robot,'' in \emph{2012 IEEE International Conference on
  Robotics and Automation}, 2012, pp. 4925--4930.

\bibitem{Struckmeier2019}
O.~{Struckmeier}, K.~{Tiwari}, M.~{Salman}, M.~J. {Pearson}, and V.~{Kyrki},
  ``Vita-slam: A bio-inspired visuo-tactile slam for navigation while
  interacting with aliased environments,'' in \emph{2019 IEEE International
  Conference on Cyborg and Bionic Systems (CBS)}, 2019, pp. 97--103.

\bibitem{Jacobson2012}
A.~Jacobson and M.~Milford, ``Towards brain-based sensor fusion for navigating
  robots,'' 01 2012.

\bibitem{Glover2019}
A.~J. {Glover}, W.~P. {Maddern}, M.~J. {Milford}, and G.~F. {Wyeth}, ``Fab-map
  + ratslam: Appearance-based slam for multiple times of day,'' in \emph{2010
  IEEE International Conference on Robotics and Automation}, 2010, pp.
  3507--3512.

\bibitem{Milford2010}
\BIBentryALTinterwordspacing
M.~Milford and G.~Wyeth, ``Persistent navigation and mapping using a
  biologically inspired slam system,'' \emph{The International Journal of
  Robotics Research}, vol.~29, no.~9, pp. 1131--1153, 2010. [Online].
  Available: \url{https://doi.org/10.1177/0278364909340592}
\BIBentrySTDinterwordspacing

\bibitem{muller2014}
S.~M{\"u}ller, C.~Weber, and S.~Wermter, ``Ratslam on humanoids - a
  bio-inspired slam model adapted to a humanoid robot,'' in \emph{Artificial
  Neural Networks and Machine Learning -- ICANN 2014}, 2014, pp. 789--796.

\bibitem{Yu2020}
\BIBentryALTinterwordspacing
S.~Yu, J.~Wu, H.~Xu, R.~Sun, and L.~Sun, ``Robustness improvement of visual
  templates matching based on frequency-tuned model in ratslam,''
  \emph{Frontiers in Neurorobotics}, vol.~14, p.~68, 2020. [Online]. Available:
  \url{https://www.frontiersin.org/article/10.3389/fnbot.2020.568091}
\BIBentrySTDinterwordspacing

\bibitem{Chen2017}
Z.~{Chen}, A.~{Jacobson}, N.~{Sünderhauf}, B.~{Upcroft}, L.~{Liu}, C.~{Shen},
  I.~{Reid}, and M.~{Milford}, ``Deep learning features at scale for visual
  place recognition,'' in \emph{2017 IEEE International Conference on Robotics
  and Automation (ICRA)}, 2017, pp. 3223--3230.

\bibitem{Kornblith2019}
S.~{Kornblith}, J.~{Shlens}, and Q.~V. {Le}, ``Do better imagenet models
  transfer better?'' in \emph{2019 IEEE/CVF Conference on Computer Vision and
  Pattern Recognition (CVPR)}, 2019, pp. 2656--2666.

\bibitem{Friston2013life}
K.~J. Friston, ``Life as we know it,'' \emph{Journal of the Royal Society
  Interface}, 2013.

\bibitem{Milford2016}
\BIBentryALTinterwordspacing
M.~Milford, A.~Jacobson, Z.~Chen, and G.~Wyeth, ``{RatSLAM: Using Models of
  Rodent Hippocampus for Robot Navigation and Beyond},'' in \emph{Robotics
  Research: The 16th International Symposium ISRR}, vol. 114, no. April, 2016,
  pp. 467--485. [Online]. Available:
  \url{http://link.springer.com/10.1007/978-3-319-28872-7}
\BIBentrySTDinterwordspacing

\bibitem{Battaglia1998}
\BIBentryALTinterwordspacing
F.~P. Battaglia and A.~Treves, ``Attractor neural networks storing multiple
  space representations: A model for hippocampal place fields,'' \emph{Phys.
  Rev. E}, vol.~58, pp. 7738--7753, Dec 1998. [Online]. Available:
  \url{https://link.aps.org/doi/10.1103/PhysRevE.58.7738}
\BIBentrySTDinterwordspacing

\bibitem{Choe2013}
\BIBentryALTinterwordspacing
Y.~Choe, \emph{Hebbian Learning}.\hskip 1em plus 0.5em minus 0.4em\relax New
  York, NY: Springer New York, 2013, pp. 1--5. [Online]. Available:
  \url{https://doi.org/10.1007/978-1-4614-7320-6\_672-1}
\BIBentrySTDinterwordspacing

\bibitem{Kingma2014}
\BIBentryALTinterwordspacing
D.~P. Kingma and M.~Welling, ``Auto-encoding variational bayes,'' in \emph{2nd
  International Conference on Learning Representations}, 2014. [Online].
  Available: \url{http://arxiv.org/abs/1312.6114}
\BIBentrySTDinterwordspacing

\bibitem{catal2020frontiers}
O.~{\c{C}atal}, S.~{Wauthier}, C.~{De Boom}, T.~{Verbelen}, and B.~{Dhoedt},
  ``Learning generative state space models for active inference,''
  \emph{Frontiers in Computational Neuroscience}, 2020, in Press.

\bibitem{schmidhuber1997}
\BIBentryALTinterwordspacing
S.~Hochreiter and J.~Schmidhuber, ``Long short-term memory,'' \emph{Neural
  Computation}, vol.~9, no.~8, pp. 1735--1780, 1997. [Online]. Available:
  \url{https://doi.org/10.1162/neco.1997.9.8.1735}
\BIBentrySTDinterwordspacing

\bibitem{dugas2000}
C.~Dugas, Y.~Bengio, F.~Bélisle, C.~Nadeau, and R.~Garcia, ``Incorporating
  second-order functional knowledge for better option pricing.'' in
  \emph{Proceedings of the 13th International Conference on Neural Information
  Processing Systems NIPS'00}.\hskip 1em plus 0.5em minus 0.4em\relax MIT
  Press, 01 2000, pp. 472--478.

\bibitem{Maas2013RectifierNI}
A.~L. Maas, ``Rectifier nonlinearities improve neural network acoustic
  models,'' in \emph{Proceedings of the 30th International Conference on
  Machine Learning}, 2013.

\bibitem{Kingma14adam}
\BIBentryALTinterwordspacing
D.~P. Kingma and J.~Ba, ``Adam: {A} method for stochastic optimization,'' in
  \emph{3rd International Conference on Learning Representations, {ICLR} 2015,
  San Diego, CA, USA, May 7-9, 2015, Conference Track Proceedings}, Y.~Bengio
  and Y.~LeCun, Eds., 2015. [Online]. Available:
  \url{http://arxiv.org/abs/1412.6980}
\BIBentrySTDinterwordspacing

\bibitem{Rezende2018GeneralizedEW}
D.~J. Rezende and F.~Viola, ``Generalized elbo with constrained optimization ,
  geco,'' in \emph{Proceedings of the Workshop Bayesian Deep Learning
  (NIPS2018)}, 2018.

\bibitem{Labbe2019}
\BIBentryALTinterwordspacing
M.~Labbé and F.~Michaud, ``Rtab-map as an open-source lidar and visual
  simultaneous localization and mapping library for large-scale and long-term
  online operation,'' \emph{Journal of Field Robotics}, vol.~36, no.~2, pp.
  416--446, 2019. [Online]. Available:
  \url{https://onlinelibrary.wiley.com/doi/abs/10.1002/rob.21831}
\BIBentrySTDinterwordspacing

\bibitem{Jeannerod2003}
M.~Jeannerod, \emph{Action Monitoring and Forward Control of Movements}.\hskip
  1em plus 0.5em minus 0.4em\relax MIT Press, 2003, pp. 83--85.

\bibitem{He2016}
K.~He, X.~Zhang, S.~Ren, and J.~Sun, ``Deep residual learning for image
  recognition,'' in \emph{Proceedings of the IEEE Conference on Computer Vision
  and Pattern Recognition (CVPR)}, June 2016.

\bibitem{catal2020iros}
O.~\c{C}atal, S.~Leroux, C.~De~Boom, T.~Verbelen, and B.~Dhoedt, ``Anomaly
  detection for autonomous guided vehicles using bayesian surprise,'' in
  \emph{2020 IEEE/RSJ International Conference on Intelligent Robots and
  Systems (IROS)}, 10 2020.

\bibitem{catal2020icassp}
O.~{\c{C}atal}, T.~{Verbelen}, J.~{Nauta}, C.~D. {Boom}, and B.~{Dhoedt},
  ``Learning perception and planning with deep active inference,'' in
  \emph{ICASSP 2020 - 2020 IEEE International Conference on Acoustics, Speech
  and Signal Processing (ICASSP)}, 2020, pp. 3952--3956.

\end{thebibliography}

\end{document}